\documentclass[pdflatex,sn-mathphys,iicol]{sn-jnl}

\jyear{2021}%

\theoremstyle{thmstyleone}%
\theoremstyle{thmstyletwo}%

\theoremstyle{thmstylethree}%

\usepackage{graphicx}
\usepackage{subfigure}
\usepackage{amsmath}
\usepackage{amssymb}
\usepackage{dsfont}
\usepackage{bm}
\usepackage{booktabs}
\usepackage{wrapfig}
\usepackage{multirow}
\usepackage{url}

\usepackage{enumitem}

\usepackage{color}

\begin{document}

\title[Article Title]{RandoMix: A Mixed Sample Data Augmentation Method with Multiple Mixed Modes}


\author[1,2]{\fnm{Xiaoliang} \sur{Liu}}\email{xiaoliang\_liu@smail.nju.edu.cn}

\author*[1,3]{\fnm{Furao} \sur{Shen}}\email{frshen@nju.edu.cn}
\author[4]{\fnm{Jian} \sur{Zhao}}\email{jianzhao@nju.edu.cn}
\author[1,2]{\fnm{Changhai} \sur{Nie}}\email{changhainie@nju.edu.cn}

\affil[1]{\orgdiv{State Key Laboratory for Novel Software Technology},  \orgname{Nanjing University}, \orgaddress{\country{China}}}

\affil[2]{\orgdiv{Department of Computer Science and Technology}, \orgname{Nanjing University}, \orgaddress{\country{China}}}
\affil[3]{\orgdiv{School of Artificial Intelligence}, \orgname{Nanjing University}, \orgaddress{\country{China}}}
\affil[4]{\orgdiv{School of Electronic Science and Engineering}, \orgname{Nanjing University}, \orgaddress{\country{China}}}


\abstract{Data augmentation plays a crucial role in enhancing the robustness and performance of machine learning models across various domains. In this study, we introduce a novel mixed-sample data augmentation method called RandoMix. RandoMix is specifically designed to simultaneously address robustness and diversity challenges. It leverages a combination of linear and mask mixed modes, introducing flexibility in candidate selection and weight adjustments. We evaluate the effectiveness of RandoMix on diverse datasets, including CIFAR-10/100, Tiny-ImageNet, ImageNet, and Google Speech Commands. Our results demonstrate its superior performance compared to existing techniques such as Mixup, CutMix, Fmix, and ResizeMix. Notably, RandoMix excels in enhancing model robustness against adversarial noise, natural noise, and sample occlusion. The comprehensive experimental results and insights into parameter tuning underscore the potential of RandoMix as a versatile and effective data augmentation method. Moreover, it seamlessly integrates into the training pipeline.
}

\maketitle


\section{Introduction}
\label{sec:intro}

Deep neural networks have demonstrated remarkable success in diverse artificial intelligence tasks, including computer vision, natural language processing, speech recognition, and signal processing \cite{voulodimos2018deep, kamath2019deep}. The key factor contributing to the effectiveness of deep neural networks is the substantial number of learnable parameters. However, insights from the Vapnik-Chervonenkis (VC) theory~\cite{vapnik1968uniform} suggest that in scenarios with limited or insufficient training data, an abundance of learnable parameters increases the risk of overfitting to the training data. Consequently, the model's generalization capacity to data beyond the training distribution becomes severely constrained. To mitigate overfitting and enhance the generalization ability of neural networks, practitioners often resort to data augmentation techniques.

In recent years, a range of mixed-sample data augmentation methods~\cite{mixup, cutmix, qin2020resizemix, harris2020fmix, puzzlemix, kim2021comixup} has emerged, gaining widespread adoption in deep neural network training. Unlike traditional data augmentation, which typically considers the vicinity of samples within the same class, mixed-sample data augmentation extends this concept to the relationship between samples across different classes. Seminal works like Mixup~\cite{mixup} employ linear interpolation to blend training samples, while CutMix~\cite{cutmix} introduces diversity by pasting a patch from one image onto another instead of relying on interpolation. Notably, recent advancements, including SaliencyMix~\cite{uddin2021saliencymix}, Puzzle Mix~\cite{puzzlemix}, and Co-Mixup~\cite{kim2021comixup}, have concentrated on leveraging image saliency analysis to enhance mixed-sample data augmentation. However, integrating salient information into the augmentation process necessitates additional computational overhead.

Diverging from methods centered around saliency analysis for performance enhancement, our approach focuses on improving neural network performance by augmenting the diversity of mixed samples. Building upon prior methodologies~\cite{mixup, cutmix, qin2020resizemix, harris2020fmix}, we introduce RandoMix, a technique that enhances diversity and exhibits superior performance while maintaining model robustness and usability.

We evaluate the efficacy of our proposed method on multiple datasets, including CIFAR-10/100~\cite{krizhevsky2009learning}, Tiny-ImageNet~\cite{chrabaszcz2017downsampled}, ImageNet~\cite{russakovsky2015imagenet}, and Google Speech Commands~\cite{speechcommands}. Our experiments demonstrate that RandoMix outperforms other state-of-the-art mixed-sample data augmentation methods. In addition to assessing generalization performance, our robustness experiments illustrate that incorporating RandoMix during training concurrently enhances the model's resilience to adversarial noise, natural noise, and sample occlusion. Our contributions can be summarized as follows:
\begin{itemize}
    \item \textbf{Novel Data Augmentation:} We introduce RandoMix, a novel data augmentation technique that enhances both robustness and diversity in neural network training.
    \item \textbf{Superior Performance:} Through extensive experiments on various datasets, RandoMix outperforms existing data augmentation methods, improving model robustness against adversarial noise, natural noise, and sample occlusion.
    \item \textbf{Customizable and Easy Integration:} RandoMix offers customization options and can be seamlessly integrated into training pipelines.
    \item \textbf{Insights:} We provide valuable insights into parameter tuning, making our method informative for researchers and practitioners.
\end{itemize}

\section{Related Works}
\label{sec:related_works}

This section offers a comprehensive overview of related works in the realm of mixed-sample data augmentation, a pivotal element in enhancing the performance, robustness, and generalization of deep learning models. We explore several influential techniques that have made substantial contributions in this domain.

Mixed-sample data augmentation is a widely employed technique in deep learning. It involves creating new training data by blending different data samples to enhance the model's generalization and robustness. These methods have played a pivotal role in enhancing the robustness, generalization, and performance of deep neural networks.

Among these methods, Mixup~\cite{mixup} stands out as one of the earliest and most influential mixed-sample data augmentation techniques. It achieves this by combining two original data samples and their associated labels using linear interpolation. Mixup effectively promotes the learning of robust and generalized features by smoothing the decision boundaries between different classes.

CutMix~\cite{cutmix}, another prominent technique, focuses on region replacement. It augments training data by replacing a rectangular region in one image with a region from another image, contributing to enhanced model diversity.

SaliencyMix~\cite{uddin2021saliencymix}, on the other hand, leverages the saliency information within images to determine which regions to mix. This approach increases the diversity of training data and aligns more closely with human visual perception, thus improving the model's understanding of image features.

Fmix~\cite{harris2020fmix} introduces a unique approach by employing Fourier transform to generate binary masks in the frequency domain, which results in smooth mixing of two images in the spatial domain. This technique is particularly effective in tasks where spatial relationships are crucial.

Puzzle Mix~\cite{puzzlemix} takes inspiration from jigsaw puzzles, randomly swapping fragments between two images and reconstructing them. This novel approach introduces diversity by altering the spatial arrangement of features, which can be beneficial for various computer vision tasks.

Co-Mixup~\cite{kim2021comixup}, as its name suggests, combines the strengths of Mixup and CutMix. It achieves diversity by mixing two data samples using both linear interpolation and region replacement, providing a versatile set of augmented data for training.

Resizemix~\cite{qin2020resizemix} takes a size-based approach, adjusting image dimensions to encourage models to adapt to changes in object size and proportions. This technique is particularly useful in tasks where object scaling is a significant factor influencing model performance.

Manifold Mixup~\cite{manifoldmixup} is a data augmentation method that interpolates hidden states to avoid the manifold intrusion problem. 

PatchUp~\cite{faramarzi2022patchup} is a hidden state block-level regularization technique for Convolutional Neural Networks (CNNs), that is applied on selected contiguous blocks of feature maps from a random pair of samples.

The continuous evolution and diversification of mixed-sample data augmentation techniques highlight their importance in training deep neural networks. These techniques empower models to achieve superior generalization, robustness, and performance across a wide spectrum of applications.

\begin{figure*}[]
	\centering
	\includegraphics[width=0.75\linewidth]{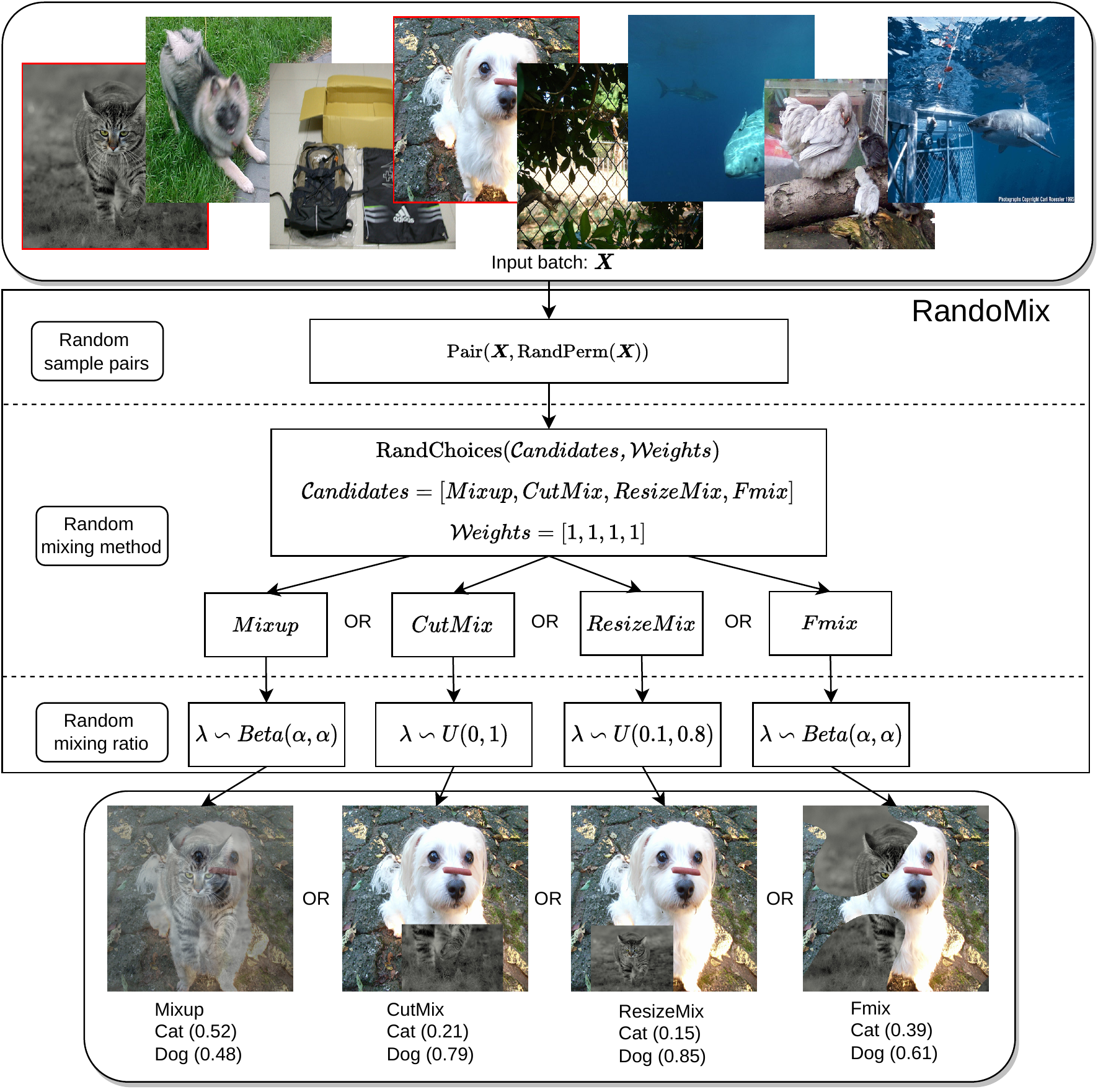}
	\caption{Illustration of the RandoMix Method. This diagram visually outlines the key steps and operations involved in the RandoMix data augmentation technique, emphasizing the random sample pairing, mixing method selection, and the subsequent impact on model training.}
	\label{fig:example}
\end{figure*}

\section{Method}
\label{sec:method}
In this section, we delve into the details of our proposed RandoMix method, a comprehensive approach designed to enhance model robustness and enrich the diversity of training data.
\subsection{Preliminaries}
Mixup~\cite{mixup} was the pioneering mixed-sample data augmentation method. It leverages the foundational principle that linear interpolations of input feature vectors should result in linear interpolations of their associated targets. In the context of Mixup, the mixing operation can be succinctly described as follows:

\begin{equation}
\begin{aligned}
\tilde{\boldsymbol{x}} &= \lambda \boldsymbol{x}_a + (1 - \lambda) \boldsymbol{x}_b,\\
\tilde{y} &= \lambda y_a + (1 - \lambda) y_b,
\end{aligned}
\end{equation}
where $(\boldsymbol{x}_a, y_a)$ and $(\boldsymbol{x}_b, y_b)$ represent two training samples, and $(\tilde{\boldsymbol{x}}, \tilde{y})$ corresponds to the generated training sample. The mixing ratio $\lambda$ is sampled from a beta distribution, specifically, $\lambda \sim Beta(\alpha, \alpha)$, with $\alpha$ falling within the interval $(0, \infty)$.

CutMix\cite{cutmix}, inspired by the concepts of Mixup and Cutout\cite{devries2017improved}, introduces a unique data augmentation strategy. It involves cutting patches from one training image and pasting them onto another, simultaneously mixing ground truth labels in proportion with the patch area. The mixing operation is mathematically defined as:

\begin{equation}
\begin{aligned}
\tilde{\boldsymbol{x}} & = \mathbf{M} \odot \boldsymbol{x}_a + (\mathbf{1}- \mathbf{M}) \odot \boldsymbol{x}_b \\
\tilde{y} & = \lambda y_a + (1-\lambda) y_b, 
\end{aligned}
\label{eq:cutmix}
\end{equation}
where $\mathbf{M}$ denotes a binary rectangular mask specifying where data is to be omitted and inserted from two samples, $\mathbf{1}$ is a binary mask filled with ones, and $\odot$ signifies element-wise multiplication. The mixing ratio $\lambda$ is sampled from a uniform distribution, specifically, $\lambda \sim \text{U}(0, 1)$.

To address challenges related to label misallocation and missing object information encountered in CutMix, ResizeMix~\cite{qin2020resizemix} employs an innovative approach. It mixes training data by resizing the source image to a smaller patch and subsequently pasting it onto another image.

Fmix~\cite{harris2020fmix} takes a unique approach. It uses random binary masks obtained by applying a threshold value to low-frequency images sampled from Fourier space, thereby further improving the shape of the CutMix mixing area.

\subsection{RandoMix}
Our proposed RandoMix is designed with the primary objective of enhancing the model's robustness and enriching the diversity of training data. To achieve this, we integrate insights and techniques from prior works, including Mixup~\cite{mixup}, CutMix~\cite{cutmix}, Fmix~\cite{harris2020fmix}, and ResizeMix~\cite{qin2020resizemix}. RandoMix not only brings an improvement in model performance but also seamlessly integrates into existing training pipelines. Fig.~\ref{fig:example} provides a visual representation of our proposed method.

As illustrated in Fig.~\ref{fig:example}, our RandoMix approach begins with random sample pairing within the input batch, $\boldsymbol{X}$. The pairing operation can be defined as follows:
\begin{equation}
\operatorname{Pairs}(\boldsymbol{X}, \operatorname{RandPerm}(\boldsymbol{X})),
\end{equation}
where $\operatorname{Pairs}$ denotes the pairing function and $\operatorname{RandPerm}$ signifies a random permutation of the input batch.

To further diversify the mixed samples, we introduce a random selection process for choosing a mixing method from a predefined list of candidates. The definition of this random selection is as follows:
\begin{equation}
\operatorname{RandChoices}(\mathcal{C}andidates, \mathcal{W}eights),
\end{equation}
where, $\operatorname{RandChoices}$ denotes random choice function. $\mathcal{C}andidates$ and $\mathcal{W}eights$ are hyperparameters. For instance, the $\mathcal{C}andidates$ can be set as $[Mixup, CutMix, ResizeMix, Fmix]$, and the $\mathcal{W}eights$ can be set as $[1, 1, 1, 1]$. There exists a one-to-one correspondence between the candidates and their respective weights.

The crucial mixing ratio $\lambda$ is meticulously determined through a corresponding random sampling process. Subsequently, these diversely mixed samples are harnessed for model training, imbuing the model with superior robustness and adaptability.

\section{Experiments}
\label{sec:experiments}
In this section, we conduct a comprehensive array of experiments to assess the performance of our proposed method. We begin by detailing the \textbf{Experimental Setup} (see Section~\ref{sec:es}). Subsequently, we present a comparative analysis of the \textbf{Generalization Performance} (see Section~~\ref{sec:gp}) of our method and other methods across four different datasets. Following this, we delve into an examination of the \textbf{Robustness and Diversity} (see Section~\ref{sec:r_and_d}) aspects of our approach.
\subsection{Experimental Setup}
\label{sec:es}
\textbf{Datasets.} CIFAR-10/100~\cite{krizhevsky2009learning} comprises 60,000 color images, each of size $32\times32$, categorized into 10 and 100 classes, respectively. Tiny-ImageNet~\cite{chrabaszcz2017downsampled} includes 200 classes, with 500 training images and 50 validation images per class. ImageNet~\cite{russakovsky2015imagenet} is a large-scale image classification benchmark dataset that contains 1.2 million training images and 50,000 validation images of 1000 classes. Google Speech Commands~\cite{speechcommands} is a speech recognition dataset featuring 105,829 one-second audio clips encompassing 35 commands.

\textbf{Network Architectures.} Different network architectures are employed for each dataset. For CIFAR-10/100, we utilize PreAct-ResNet18~\cite{he2016identity} and the WideResNet-28-10~\cite{zagoruyko2016wide}. For Tiny-ImageNet, PreAct-ResNet18~\cite{he2016identity} is the architecture of choice. When working with ImageNet, we turn to ResNet-50~\cite{he2016deep}. For the Google Speech Commands dataset, we rely on PreAct-ResNet18~\cite{he2016identity} with a customized input layer designed to accommodate one-dimensional audio signals.

\textbf{Baseline Methods.} To assess the performance of our method, we conduct a comparative analysis against various state-of-the-art mixed-sample data augmentation techniques. These include Cutout~\cite{devries2017improved}, Mixup~\cite{mixup}, CutMix~\cite{cutmix}, Fmix~\cite{harris2020fmix}, ResizeMix~\cite{qin2020resizemix}, SaliencyMix~\cite{uddin2021saliencymix}, Manifold Mixup~\cite{manifoldmixup}, PatchUp~\cite{faramarzi2022patchup}, Puzzle Mix~\cite{puzzlemix}, and Co-Mixup~\cite{kim2021comixup}. We implement these methods using the same network architectures and hyperparameters as our method for fair comparison.

 \begin{table*}[!ht]
   \centering
   \caption{Top-1 test accuracy rate (\%) on CIFAR-10 and CIFAR-100 classification with PreAct-ResNet18 (PA-RS18) and WideResNet-28-10 (WRS28-10). Top: the results of the original authors. Bottom: the results of our implementation.} 
   \label{tab:cifar10_and_100_results}
   \scalebox{1.0}{
         \begin{tabular}{lcc|cc}
         \toprule
          \multirow{2}{*}{\textbf{Method}}& \multicolumn{2}{c|}{\textbf{CIFAR-10}} & \multicolumn{2}{c}{\textbf{CIFAR-100}} \\ 
          & \textbf{PA-RS18(\%)} & \textbf{WRS28-10(\%)} & \textbf{PA-RS18(\%)} & \textbf{WRS28-10(\%)} \\ 
         \midrule
         Mixup~\cite{mixup} & 95.80 & 97.30 & 78.90 & 82.50  \\
         Cutout~\cite{devries2017improved} & - & 96.92 & - & 81.59  \\
         Fmix~\cite{harris2020fmix} & 96.14 & 96.38  & 79.85 & 82.03\\
         ResizeMix~\cite{qin2020resizemix} & - & 97.60 & - & 84.31 \\
    SaliencyMix~\cite{uddin2021saliencymix} & -   & 97.24&  - & 83.44 \\
    Manifold Mixup~\cite{manifoldmixup} & 97.05 & 97.45 & 79.66 & 81.96\\
         PatchUp~\cite{faramarzi2022patchup}& 97.08 & 97.47 & 80.88 & 82.30 \\
         Puzzle Mix~\cite{puzzlemix} & - & - & - & 84.05 \\
         Co-Mixup~\cite{kim2021comixup}& - & - & 80.13 & - \\
         \midrule
         Baseline & 95.18 & 96.28& 77.55 & 80.87  \\
         Mixup~\cite{mixup} & 96.15($\uparrow$0.97) & 97.23($\uparrow$0.95)& 79.49($\uparrow$1.94) & 82.61($\uparrow$1.74)   \\
         CutMix~\cite{cutmix} & 96.43($\uparrow$1.25) & 97.06($\uparrow$0.78)  & 79.73($\uparrow$2.18) & 82.34($\uparrow$1.47)\\
         Fmix~\cite{harris2020fmix} & 96.46($\uparrow$1.28) & 97.02($\uparrow$0.74) &79.45($\uparrow$1.90)& 81.78($\uparrow$0.97)\\
         ResizeMix~\cite{qin2020resizemix} & 96.69($\uparrow$1.51) & 97.48($\uparrow$1.20) & 80.66($\uparrow$3.11) & 83.00($\uparrow$2.13)\\
         SaliencyMix~\cite{uddin2021saliencymix} &  96.28($\uparrow$1.10) & 96.87($\uparrow$0.59) &  80.01($\uparrow$2.46) & 81.95($\uparrow$1.08)  \\
         Puzzle Mix~\cite{puzzlemix} & 96.82($\uparrow$1.64) & 97.51($\uparrow$1.23)  & 80.68($\uparrow$3.13) & 83.71($\uparrow$2.84)\\
         RandoMix & \textbf{97.22}($\uparrow$\textbf{2.04}) & \textbf{98.02}($\uparrow$\textbf{1.74})& \textbf{82.14}($\uparrow$\textbf{4.59}) & \textbf{84.84}($\uparrow$\textbf{3.97})   \\
		\bottomrule
         \end{tabular}
      }
   \end{table*}

\subsection{Generalization Performance}
\label{sec:gp}
\subsubsection{Experiments on CIFAR-10 and CIFAR-100}

We conduct experiments using two different network architectures, PreAct-ResNet18~\cite{he2016identity} and WideResNet-28-10~\cite{zagoruyko2016wide}, on the CIFAR-10 dataset. The training process extends over 200 epochs with a batch size of 256. Stochastic gradient descent (SGD) is employed with an initial learning rate of 0.1 and decays following the cosine annealing schedule~\cite{2016SGDR}. We set the momentum to 0.9 and apply a weight decay of $5 \times 10^{-4}$. The parameter $\alpha$ is consistently set to 1. In the PreAct-ResNet18 experiment, we configure the $\mathcal{C}andidates$ as $[Mixup, CutMix, ResizeMix, Fmix]$, and the $\mathcal{W}eights$ as $[1,1,1,1]$. In the WideResNet-28-10 experiment, $\mathcal{C}andidates$ is set to $[Mixup, CutMix, ResizeMix, Fmix]$, and $\mathcal{W}eights$ to $[3,1,1,1]$.

The CIFAR-100 dataset, while having an identical number of images to CIFAR-10, features 100 classes, each containing 600 images. In both the PreAct-ResNet18 and WideResNet-28-10 experiments, the settings and hyperparameters remain consistent with the CIFAR-10 dataset.

Table~\ref{tab:cifar10_and_100_results} shows how our proposed RandoMix compares to other state-of-the-art mixed-sample data augmentation methods on CIFAR-10 and CIFAR-100. The upper part of the table shows results from the original authors, while the lower part shows results from our implementations. Baseline refers to the method without any mixed-sample data augmentation. Notably, our proposed RandoMix consistently outperforms other mixed-sample data augmentation methods. For example, on CIFAR-100, RandoMix achieves a performance increase of $2.23\%$ compared to Mixup, $2.5\%$ over CutMix, and $1.13\%$ higher accuracy than PuzzleMix with the WideResNet-28-10 architecture.

\begin{table*}[]
   \centering
   \caption{Top-1 and Top-5 test accuracy rate (\%) on Tiny-ImageNet and ImageNet. $\textbf{Cost} = \frac{Training\ time}{Baseline\ training\ time}$.}
   \label{tab:tinyimagenet_results}
   \scalebox{1.0}{
         \begin{tabular}{lcc|ccc}
         \toprule
          \multirow{2}{*}{\textbf{Method}}& \multicolumn{2}{c|}{\textbf{Tiny-ImageNet}} & \multicolumn{3}{c}{\textbf{ImageNet}} \\ 
          & \textbf{Top-1(\%)} & \textbf{Top-5(\%)} & \textbf{Top-1(\%)} & \textbf{Top-5(\%)} & \textbf{Cost} \\ 
         \midrule
         Baseline  & 63.43 & 83.05 & 76.37 & 93.04 & 1.00 \\ 
         Mixup~\cite{mixup}  & 64.77($\uparrow$1.34) & 84.05($\uparrow$1.00) & 77.42($\uparrow$1.05) & 93.67  ($\uparrow$0.63)& 1.00 \\ 
         CutMix~\cite{cutmix}  &67.72($\uparrow$4.29) & 86.25($\uparrow$3.20)& 77.32($\uparrow$0.95) & 93.79($\uparrow$0.75) & 0.99 \\
         Fmix~\cite{harris2020fmix} & 66.18($\uparrow$2.75) & 84.58($\uparrow$1.53)& 77.27($\uparrow$0.90) & 93.62($\uparrow$0.58) & 1.07  \\
         ResizeMix~\cite{qin2020resizemix} & 67.71($\uparrow$4.28) & 86.65({$\uparrow$3.60}) & 77.83($\uparrow$1.46) & 93.87({$\uparrow$0.83}) & 0.99 \\ 
  	     SaliencyMix~\cite{uddin2021saliencymix} & 65.90($\uparrow$2.47) & 85.13($\uparrow$2.08)& 77.40($\uparrow$1.03) & 93.69($\uparrow$0.65) & 1.01 \\
         Puzzle Mix~\cite{puzzlemix}  & 66.65($\uparrow$3.22) & 85.29({$\uparrow$2.24})& 77.69($\uparrow$1.32) & 93.94($\uparrow${0.90}) &  2.10\\
         RandoMix  &  \textbf{68.94}(\textbf{$\uparrow$5.51}) & \textbf{86.83}(\textbf{$\uparrow$3.78})&  \textbf{77.88}($\uparrow$\textbf{1.51}) & \textbf{93.98}({$\uparrow$\textbf{0.94}}) & 1.01 \\
         \bottomrule
         \end{tabular}     
      }
   \end{table*}

\subsubsection{Experiments on Tiny-ImageNet and ImageNet}
We conduct experiments using the PreAct-ResNet18 network on the Tiny-ImageNet dataset. The training spans 200 epochs with a batch size of 256. We apply the stochastic gradient descent (SGD) optimization method with an initial learning rate of 0.1. This rate decreases as we follow the cosine annealing schedule. We keep the momentum constant at 0.9 and incorporate weight decay at $5 \times 10^{-4}$. The parameters $\mathcal{C}andidates$ are specified as $[Mixup, CutMix, ResizeMix, Fmix]$, $\mathcal{W}eights$ as $[1, 1, 1, 1]$, and $\alpha$ as 1.

For the ImageNet dataset, we utilize the ResNet-50~\cite{he2016deep} architecture. Training is executed on 8 NVIDIA V100 GPUs. We apply standard data augmentation techniques, such as $224 \times 224$ random resized cropping and random horizontal flipping, following practices outlined in references such as CutMix, PuzzleMix, SaliencyMix, and Fmix. Stochastic gradient descent (SGD) with an initial learning rate of 0.8 is employed, subject to a decay regimen based on the cosine annealing schedule. A momentum value of 0.9 is maintained, and weight decay is configured at $4 \times 10^{-5}$. Training spans 100 epochs with a batch size of $256 \times 8$. The parameters $\mathcal{C}andidates$ are set as $[Mixup, CutMix, ResizeMix, Fmix]$, $\mathcal{W}eights$ as $[3, 1, 1, 1]$, and $\alpha$ as 0.2.

The results presented in Table~\ref{tab:tinyimagenet_results} demonstrate that our method consistently outperforms alternative approaches on both the Tiny-ImageNet and ImageNet datasets. Importantly, the computational "Cost" associated with our method remains comparable to the "Baseline," emphasizing its efficiency and the minimal additional computational requirements.

\begin{table}[!ht]
   \centering
   \caption{Top1 validation and test accuracy rate (\%) on  Google Speech Commands.}
   \label{tab:commands_results}
   \scalebox{0.9}{
         \begin{tabular}{lcccc}
         \toprule
         \textbf{Method}  & \textbf{Val Top-1(\%)} & \textbf{Test Top-1(\%)}  \\ 
         \midrule
         Baseline  & 97.98 & 98.25  \\ 
         Mixup~\cite{mixup}  & 98.26($\uparrow$0.28) &  98.59 ($\uparrow$0.34)\\ 
         CutMix~\cite{cutmix}  & 97.93($\downarrow$0.05) & 98.36($\uparrow$0.11)  \\
         Fmix~\cite{harris2020fmix} & 98.13($\uparrow$0.15) & 98.60($\uparrow$0.35)   \\
         ResizeMix~\cite{qin2020resizemix} & 98.14($\uparrow$0.16) & 98.53({$\uparrow$0.28}) & \\ 
  	     
         RandoMix &\textbf{98.33}($\uparrow$\textbf{0.35}) & \textbf{98.77}({$\uparrow$\textbf{0.52}})\\
         \bottomrule
         \end{tabular}
      }
   \end{table}

\subsubsection{Experiments on Google Speech Commands}
Google Speech Commands~\cite{speechcommands} constitutes a speech recognition dataset, with specific emphasis on a defined set of keywords. Following the structure established by the Kaggle challenge~\cite{kagglespeechcommands}, our experimental focus revolves around the recognition of ten predefined classes: "yes," "no," "up," "down," "left," "right," "on," "off," "stop," and "go." Any other utterances are categorized as "unknown" or "silence."

Our experiments entail training the PreAct-ResNet18 network over 300 epochs, utilizing a batch size of 256 and employing the stochastic gradient descent (SGD) optimizer. The training configuration involves an initial learning rate of 0.05, weight decay set at $5 \times 10^{-4}$, and a momentum value of 0.9. For this task, we set the parameter $\alpha$ to 1.0, while $\mathcal{C}andidates$ are denoted as $[Mixup, CutMix, ResizeMix, Fmix]$, with $\mathcal{W}eights$ assigned as $[1, 1, 1, 1]$.

The results pertaining to Google Speech Commands are comprehensively detailed in Table~\ref{tab:commands_results}. Notably, unlike Puzzle Mix~\cite{puzzlemix} and SaliencyMix~\cite{uddin2021saliencymix}, which rely on image saliency analysis and are inherently challenging to adapt to speech recognition tasks, our method offers a versatile and adaptable solution for speech recognition applications. Significantly, it attains the most promising results in this context.

\begin{table*}[!ht]
   \centering
   \caption{Robustness experimental results on CIFAR-100 with WideResNet-28-10. In \textbf{Mixed Mode}, ``$l$'' denotes linear mixed mode and ``$m$'' denotes mask mixed mode.}
   \label{tab:robustness}
   \scalebox{1.0}{
         \begin{tabular}{lc|c|cc|c}
         \toprule
         \multirow{2}{*}{\textbf{Method}}&\textbf{Mixed}& \textbf{Clean} &\textbf{FGSM} & \textbf{Corruption} & \textbf{Occlusion}  \\ 
         & \textbf{Mode}&\textbf{Error(\%)} &\textbf{Error(\%)} & \textbf{Error(\%)} & \textbf{Error(\%)}  \\
         \midrule
         Baseline & - & 19.27 & 84.63 & 49.42 & 42.50 \\ 
         Mixup~\cite{mixup} & $l$ & 17.49 & 64.43(\textcolor{green}{$\downarrow$}) & \textbf{41.67}(\textcolor{green}{$\downarrow$}) & 43.40(\textcolor{red}{$\uparrow$}) \\ 
         CutMix~\cite{cutmix} & $m$ & 17.46 & 84.70(\textcolor{red}{$\uparrow$}) & 52.91(\textcolor{red}{$\uparrow$}) & 33.44(\textcolor{green}{$\downarrow$})  \\
         Fmix~\cite{harris2020fmix} & $m$ & 18.64 & 87.41(\textcolor{red}{$\uparrow$}) & 51.61(\textcolor{red}{$\uparrow$}) &  28.96(\textcolor{green}{$\downarrow$}) \\
         ResizeMix~\cite{qin2020resizemix} & $m$ & 17.12 & 90.04(\textcolor{red}{$\uparrow$}) & 50.37(\textcolor{red}{$\uparrow$}) & \textbf{27.29}(\textcolor{green}{$\downarrow$}) \\ 
	\midrule
         RandoMix & $l+m$ & \textbf{15.16} & \textbf{63.78}(\textcolor{green}{$\downarrow$}) & 42.17(\textcolor{green}{$\downarrow$}) & {27.72}(\textcolor{green}{$\downarrow$}) \\
         \bottomrule
         \end{tabular}
      }
   \end{table*}

\begin{table*}[!ht]
   \centering
   \caption{The relationship between parameter $\mathcal{W}eights$ and robustness.}
   \label{tab:weights}
   \scalebox{1.0}{
         \begin{tabular}{lc|c|cc|c}
         \toprule
         \multirow{2}{*}{\textbf{Method}}&\multirow{2}{*}{\textbf{$\mathcal{W}eights$}} & \textbf{Clean} &\textbf{FGSM} & \textbf{Corruption} & \textbf{Occlusion}  \\ 
         & \textbf{}&\textbf{Error(\%)} &\textbf{Error(\%)} & \textbf{Error(\%)} & \textbf{Error(\%)}  \\
         \midrule
         RandoMix & [1,1,1,1] & 15.35 & 75.81 & 45.84 & \textbf{26.73} \\
         RandoMix & [2,1,1,1] & 15.65 & 66.64 & 42.45 & 29.55 \\
         RandoMix & [3,1,1,1] & 15.51 & 70.23 & \textbf{42.16} & 28.12\\
         RandoMix & [4,1,1,1] & \textbf{15.16}  & \textbf{63.78} & {42.17} & 27.72\\
         \bottomrule
         \end{tabular}
      }
   \end{table*}

\subsection{Robustness and Diversity}
\label{sec:r_and_d}
\subsubsection{Robustness and $\mathcal{W}eights$}
In our robustness experiments, we employ the WideResNet-28-10 model on CIFAR-100 and assess its performance against common robustness challenges, including adversarial noise, natural noise, and sample occlusion.

For adversarial noise robustness assessment, we utilize the Fast Gradient Sign Method (FGSM)~\cite{goodfellow2014explaining} attack with an $\ell_\infty$ epsilon ball of $8/255$. To evaluate natural noise robustness, we employ CIFAR-100-C~\cite{hendrycks2019benchmarking} and, following established practices~\cite{hendrycks2019benchmarking}, utilize the mean Corruption Error as the metric. In the case of sample occlusion robustness, we generate occluded samples by introducing random occlusion blocks filled with zeros.

The robustness experimental results are summarized in Table~\ref{tab:robustness}. We categorize Mixup~\cite{mixup}, CutMix~\cite{cutmix}, Fmix~\cite{harris2020fmix}, and ResizeMix~\cite{qin2020resizemix} into two groups based on their mixed modes, namely linear and mask mixed modes. As shown in Table~\ref{tab:robustness}, methods based on the linear mixed mode exhibit superior noise robustness but lower occlusion robustness. Conversely, methods based on the mask mixed mode demonstrate the opposite pattern. In comparison to single mixed mode methods, our approach (RandoMix) leverages both linear and mask mixed modes, enhancing both noise and occlusion robustness. Notably, it achieves the lowest Clean Error. The connection between the $\mathcal{W}eights$ parameter and robustness is illustrated in Table~\ref{tab:weights}. Although the relationship between $\mathcal{W}eights$ and robustness lacks linearity due to inherent randomness, we can still fine-tune the robustness against noise and occlusion by adjusting the $\mathcal{W}eights$.

\begin{table}[!ht]
   \centering
   \caption{Diversity experiment results on Tiny-ImageNet with PreAct-ResNet18. In \textbf{$\mathcal{C}andidates$}, ``${M}$'' denotes Mixup, ``${C}$'' denotes CutMix, ``${R}$'' denotes ResizeMix, and ``${F}$'' denotes Fmix. In \textbf{Mixed Mode}, ``$l$'' denotes linear mixed mode and ``$m$'' denotes mask mixed mode. In all candidate combinations, the $\mathcal{W}eights$ of all candidates are equal.} 
   \label{tab:candidates}
   \scalebox{0.9}{
         \begin{tabular}{lcc|c}
         \toprule
          \textbf{$\mathcal{C}andidates$}& \textbf{Mixed Mode} & {\textbf{Top1(\%)}} & {\textbf{Mean(\%)}} \\ 
         \midrule
         Baseline & - & 63.75 &  63.75  \\
         \midrule
         $[{M}]$& $l$ & 65.73 & \multirow{4}{*}{66.47} \\
         $[{C}]$& $m$ & 67.17 & ~  \\
         $[{R}]$& $m$ & 67.24 & ~  \\
         $[{F}]$& $m$ & 65.77 & ~  \\
         \midrule
         $[{M,C}]$& $l+m$ &  68.50 & \multirow{5}{*}{67.97}\\
         $[{M,R}]$& $l+m$ &  68.89 & \\
         $[{M,F}]$& $l+m$ &  67.81 & \\
         $[{C,R}]$& $m$ &  67.73 & \\
         $[{C,F}]$& $m$ &  66.94 & \\
		 \midrule	
         $[{M,C,R}]$& $l+m$ & 68.91  &\multirow{4}{*}{68.68}  \\
         $[{M,C,F}]$& $l+m$ &  68.56 & \\
         $[{M,R,F}]$& $l+m$ & 	\textbf{69.07} & \\
         $[{C,R,F}]$& $m$ & 68.19  & \\
         \midrule
         $[{M,C,R,F}]$& $l+m$ & {68.92} & \textbf{68.92}\\
         
         \bottomrule
         \end{tabular}
      }
   \end{table}

\subsubsection{Diversity and $\mathcal{C}andidates$}
Not only does our method excel in robustness, but it also shines in terms of diversity. The diversity experiment results, conducted on Tiny-ImageNet with the PreAct-ResNet18 model, are presented in Table~\ref{tab:candidates}. These results illustrate that, when considering an equal number of $\mathcal{C}andidates$, configurations that incorporate both linear and mask mixed modes outperform those utilizing a single mixed mode exclusively. Additionally, the table reveals that, with a greater number of candidates, RandoMix exhibits improved performance, as indicated by the "Mean" metric.

As showcased in Table~\ref{tab:candidates}, our method, in contrast to prior approaches~\cite{mixup,cutmix,harris2020fmix,qin2020resizemix}, offers an enhanced capacity to diversify the training data by adjusting the $\mathcal{C}andidates$ parameter, thereby yielding superior performance.

\section{Discussion}
\label{sec:discussion}
The results presented in the previous sections demonstrate the effectiveness of our proposed method, RandoMix, in enhancing both robustness and diversity in various experimental settings. In this section, we delve into the implications and broader context of these findings, discussing the key takeaways and potential avenues for further research.

\textbf{Advancements in Data Augmentation:}
RandoMix showcases remarkable improvements over existing data augmentation techniques such as Mixup, CutMix, Fmix, and ResizeMix. By integrating both linear and mask mixed modes and allowing for the adjustment of $\mathcal{C}andidates$, our approach significantly bolsters the diversity of the training data, which is a crucial factor in improving model performance. Moreover, the adaptability of our method to different datasets and tasks, as evidenced by our experiments on CIFAR-10/100, Tiny-ImageNet, ImageNet, and Google Speech Commands, underscores its versatility and practicality.

\textbf{Robustness Across Various Challenges:}
Robustness is a pivotal aspect of model reliability. Our experiments on CIFAR-100 showcased the superiority of RandoMix in addressing challenges related to adversarial noise, natural noise, and sample occlusion. The incorporation of both linear and mask mixed modes permits RandoMix to effectively handle adversarial noise and occlusion simultaneously. Furthermore, by examining the relationship between the $\mathcal{W}eights$ and robustness, we gain insights into how the fine-tuning of parameters can further enhance robustness, even in the presence of randomness.

\textbf{Implications for Future Research:}
The success of RandoMix raises questions about the full potential of mixed-sample data augmentation techniques. Further exploration into the nuanced interactions between mixed modes, candidate selection, and hyperparameters could yield even more robust and diverse training strategies. Additionally, extending the application of RandoMix to more complex domains such as natural language processing or reinforcement learning may uncover new dimensions of its capabilities.

\section{Conclusion}
\label{sec:conclusion}
In this conclusion, we introduced RandoMix, a versatile mixed-sample data augmentation method with multiple mixed modes. Our experiments have consistently demonstrated that RandoMix outperforms other mixed-sample data augmentation techniques in the context of image classification and speech recognition tasks.

Furthermore, a significant revelation from our research is that single mixed modes cannot simultaneously enhance a model's robustness against various forms of challenges, such as noise and occlusion. However, RandoMix distinguishes itself by uniquely offering concurrent improvements in model robustness against adversarial noise, natural noise, and sample occlusion. This multifaceted approach opens up new possibilities in enhancing model resilience under diverse conditions.

In our future endeavors, we plan to delve deeper into the application of mixed-sample data augmentation in multi-modal tasks. Additionally, we will explore the profound impact of mixed-sample data augmentation on model robustness, seeking to uncover further insights that can empower the field of deep learning.

\section*{Acknowledgments}
This work was supported in part by the STI 2030-Major Projects of China under Grant 2021ZD0201300, and by the National Science Foundation of China under Grant 62276127.

\section*{Declarations}
\noindent\textbf{Data Availability.}
The data that support the findings of this study are available from the corresponding author upon reasonable request.

\noindent\textbf{Conflict of Interest.} The authors declare that they have no conflict of interest.

\bibliography{references}

\vspace{1cm}
\end{document}